%% file: PNAS_commentary.tex
\documentclass[11pt,a4paper]{article}
\usepackage[utf8]{inputenc}
\usepackage{geometry}
\usepackage{hyperref}
\usepackage{graphicx}
\usepackage{tikz}
\usetikzlibrary{arrows.meta, positioning, calc}
\usepackage{caption}

\title{\textbf{The Self-Assembling Brain: Emergent Routing and Its Open Challenges}}
\author{Dongsung Huh$^{1, 2}$}
\date{
    \small $^{1}$The NSF Institute for Artificial Intelligence and Fundamental Interactions (IAIFI) \\
    \small $^{2}$Centaur AI Institute 
    \vspace{1em} 
}

\begin{document}

\maketitle

Artificial neural networks have become powerful models of sensory processing and cognition, yet how the brain might implement their learning algorithms remains a central mystery. In machine learning, deep networks learn via backpropagation, which requires that the synaptic weights carrying error signals backward precisely mirror the weights carrying sensory signals forward. Backpropagation also assumes that the backward pathway knows the exact local excitability---the mathematical derivative---of every neuron in the forward pass. Both requirements sit uneasily with cortical neuroanatomy.

Over the past decade, biologically inspired alternatives such as feedback alignment have shown that networks can still learn when exact weight symmetry is relaxed \cite{Lillicrap2016}. Yet even these models leave a deeper developmental puzzle untouched: standard feedback-alignment formulations presuppose a fixed correspondence between forward and feedback streams at each layer, with forward nonlinearities embedded directly into the feedback pass. How such reciprocal organization could arise developmentally without being specified in advance remains an open problem. In this issue of PNAS, Liao et al. call this impasse ``architectural circularity'': existing models explain how a fixed circuit learns, while presupposing the very learning circuit whose biological origin is in question \cite{Liao2026}.

\section*{Building a Learning Circuit from Local Rules}

Liao et al. reframe biological plausibility not as a binary checklist of forbidden machine-learning operations, but as a graded axis of self-organization. The further a model moves along this axis, the less wiring it stipulates in advance and the more it derives from local synaptic interactions. To demonstrate how a credit-assignment circuit---the machinery that determines which synapses should change during learning---might emerge, the authors introduce a minimal building block called the Self-Assembling Motif (SAM).

SAM begins with a spare anatomical scaffold comprising two neuronal streams: an ascending sensory-inference stream and a descending error-feedback stream, linked by cross-stream connections. Within each stream, synapses act as traditional driving connections that propagate neural activity from layer to layer. Between the two streams, however, connections are purely modulatory; they do not drive firing in the opposing stream, but instead deliver teaching signals that guide local synaptic plasticity. Crucially, unlike prior models, SAM does not wire the two streams together in a fixed one-to-one correspondence. The cross-stream bridges start out random, unassigned, and fully plastic, and the two streams need not even contain the same number of neurons \cite{Liao2026}.

What makes SAM conceptually distinctive is how it separates local bridge alignment from network-level pathway alignment \cite{Liao2026} (Fig.~\ref{fig:sam_conceptual}). At the local level, under matched learning rates and the other stated assumptions, SAM's update rules guarantee alignment of the reciprocal cross-stream connections. Correlated local activity brings their weights into agreement, transforming initially tangled connectivity into coherent cross-stream bridges. Separately, when these motifs are stacked across layers, the authors observe empirically that the forward and backward pathways can become functionally aligned during stable training, such that the feedback route carries information appropriate to the forward computation. The resulting synaptic updates can then follow directions that approximate gradient descent.

Notably, the resulting learning dynamics do not simply replicate textbook stochastic gradient descent. Instead, the self-organized cross-stream connections produce \textit{emergent preconditioning}---an adaptive learning rate that evolves alongside the task and that the authors liken to biological metaplasticity. Because SAM drops fixed neuron-to-neuron pairing, the feedback stream can also be much wider than the forward stream. On the reported image-classification benchmarks, SAM approaches or, under specified conditions, exceeds the performance of the compared backpropagation baseline. In these simulations, performance peaks when feedback connections outnumber feedforward ones by roughly five- to eight-fold---an architecture-dependent finding that intriguingly parallels the dense feedback projections observed in sensory cortex \cite{Liao2026}. By turning credit-assignment routing from a pre-wired assumption into a self-organizing process, Liao et al. demonstrate that key components of gradient-based learning can emerge from local synaptic rules.

\input{tikz_fig_source.tex}

\section*{Algorithmic Trade-Offs and Scaling Frontiers}

Moving routing machinery from a fixed blueprint to an emergent property clarifies three algorithmic challenges that define the next research frontier.

First, can self-assembled feedback preserve neuron-specific nonlinear information without fixed wiring? Dropping neuron-for-neuron correspondence between the two streams introduces a direct trade-off between wiring flexibility and gradient precision. In the linear shortcut used by SAM, the backward pathway does not track whether an individual forward unit is active or silent, functioning as a \textit{straight-through approximation} that passes error signals through nonlinear nodes unaltered \cite{Bengio2013}. While this linear pass-through performs well on standard image benchmarks, it cannot capture stimulus-dependent neuronal gating, leaving open how much gradient fidelity is lost in more strongly nonlinear tasks.

Second, what controls residual routing errors as circuits grow deeper? Although local bridge alignment is guaranteed under matched learning rates, the learning rules do not directly specify how mismatch between the forward and backward pathways is minimized. When residual pathway mismatches persist, small routing errors compound as feedback travels backward across successive layers, making earlier layers increasingly vulnerable to upstream noise and explaining why deeper configurations can lose alignment and diverge \cite{Liao2026}. Quantifying how large these residual errors remain relative to the learning signal---and identifying local mechanisms that actively bound them---will be essential for scaling self-assembling circuits to deeper hierarchies.

Third, does low-dimensional routing sacrifice important learning directions? Liao et al. observe that the emergent preconditioner naturally compresses into a \textit{low-dimensional bottleneck} during training \cite{Liao2026}. While low-dimensional structure can support helpful simplicity biases, compressed routing can also suppress learning along stimulus dimensions that fall outside its active subspace, leaving those unrepresented features more exposed to background routing noise. Whether biological circuits avoid or dynamically escape such blind spots remains an open question.

\section*{Biophysical and Temporal Questions}

Translating SAM from an algorithmic motif into a cortical circuit also raises concrete cellular and timing questions. Although the authors describe the network as governed by a ``uniform'' heterosynaptic plasticity rule, the model relies on a sharp functional divide between synapse types. Within-stream synapses are traditional driving connections that govern postsynaptic firing, whereas cross-stream connections act as non-driving conduits that contribute zero driving current and exist solely to ferry learning signals. Recent experimental work has shown that cortical dendrites can process compartmentalized instructive signals during learning \cite{Francioni2026}. Even so, SAM models a functional distinction that has not yet received a direct cortical implementation: modulatory connections that route teaching signals without contributing to ordinary postsynaptic activation.

This structural divide also splits the plasticity mechanism itself. Within-stream driving updates are heterosynaptic, guided by a third-party modulatory signal \cite{Kandel1965}, whereas updates on the non-driving conduits reduce to a direct correlation between local upstream and downstream activity, operating effectively as a standard homosynaptic Hebbian rule \cite{Liao2026}. Distinguishing these two biophysical roles will be important when designing experiments to test whether cortical microcircuits implement SAM-like motifs.

Finally, SAM relies on sequential, alternating phases in which the forward stream completes inference on a static input before the backward stream propagates errors and updates synapses. Cortical processing, by contrast, combines asynchronous local activity with distributed timing mechanisms to handle an uninterrupted sensory stream---closer to continuous, unsegmented video than to isolated still frames---rather than executing a globally imposed inference-then-update alternation \cite{Bacvanski2026}. While Liao et al. reasonably propose that synaptic eligibility traces---transient local records of recent activity---could help bridge temporal gaps, reconciling turn-based updates with continuous, non-stationary sensory input remains a key challenge \cite{Liao2026, Bacvanski2026}.

\section*{Outlook}

Liao et al. take an illuminating step past one of the most persistent wiring assumptions in biologically inspired learning. By reframing credit-assignment circuitry as a self-organizing process rather than a pre-specified blueprint, SAM turns questions of nonlinear precision, depth stability, low-dimensional bottlenecks, and cortical biophysics into experimentally and theoretically testable problems. Building on this foundation, future models can connect bottom-up local plasticity rules with top-down stability principles, moving closer to a complete account of how biological circuits assemble themselves to learn.

\end{document}

%% file: tikz_fig_source.tex
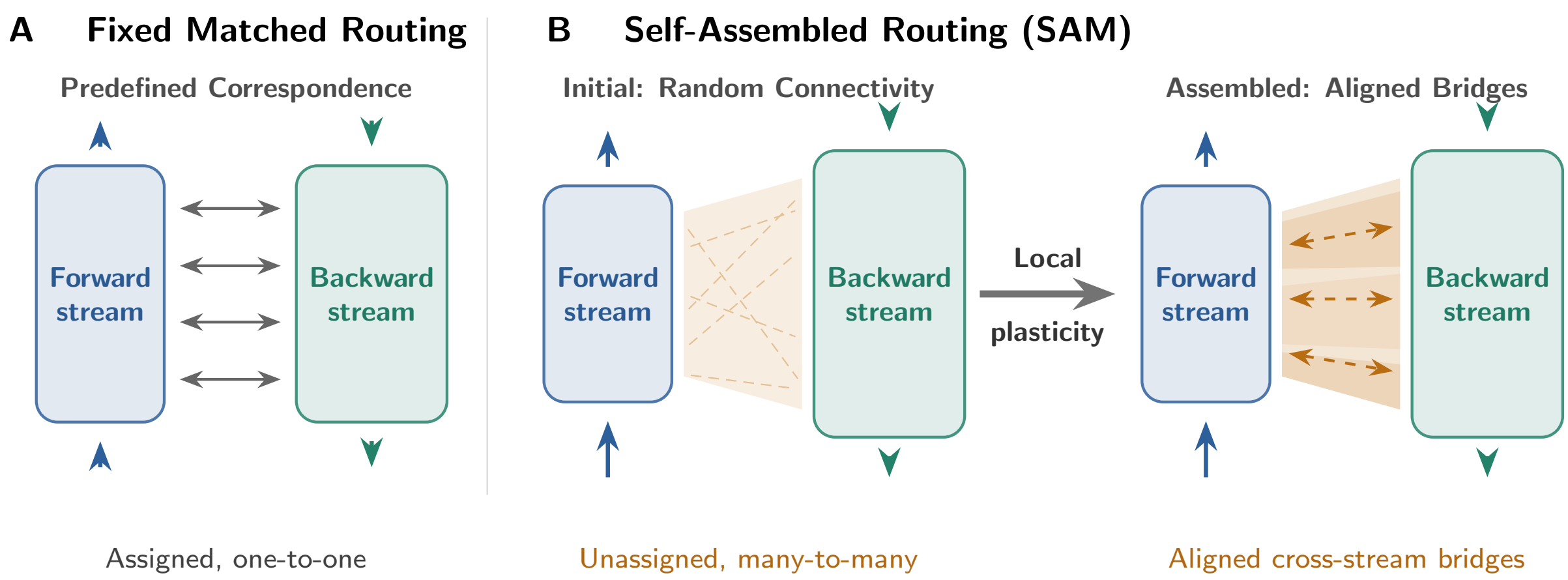
\begin{figure*}[t!]
\centering
\definecolor{fwdCol}{RGB}{45, 95, 155}
\definecolor{bwdCol}{RGB}{35, 130, 105}
\definecolor{modCol}{RGB}{195, 115, 20}

\resizebox{\textwidth}{!}{%
\begin{tikzpicture}[
    font=\sffamily,
    >=Stealth,
    popBlock/.style={
        rectangle, rounded corners=6pt, draw=#1!85, fill=#1!14,
        thick, minimum width=1.15cm, align=center, font=\scriptsize\sffamily\bfseries, text=#1!95!black
    },
    driveArrow/.style={->, very thick, #1, shorten <=5pt, shorten >=5pt},
    fixedLine/.style={<->, thick, black!60, shorten <=2pt, shorten >=2pt}
]

% =========================================================================
% PANEL A: Fixed Matched Routing (Prior Models)
% =========================================================================
\begin{scope}[xshift=0cm]
    \node[anchor=west, font=\small\sffamily\bfseries] at (0.0, 4.95)
        {A \quad Fixed Matched Routing};

    % Equal-size population blocks
    \node[popBlock=fwdCol, minimum height=2.55cm] (A_fwd) at (1.05, 2.35)
        {Forward\\stream};
    \node[popBlock=bwdCol, minimum height=2.55cm] (A_bwd) at (3.75, 2.35)
        {Backward\\stream};

    % Driving within-stream arrows
    \draw[driveArrow=fwdCol] (1.05, 0.45) -- (A_fwd.south);
    \draw[driveArrow=fwdCol] (A_fwd.north) -- (1.05, 4.25);
    \draw[driveArrow=bwdCol] (3.75, 4.25) -- (A_bwd.north);
    \draw[driveArrow=bwdCol] (A_bwd.south) -- (3.75, 0.45);

    % Fixed paired links
    \foreach \yf/\yb in {3.20/3.20, 2.63/2.63, 2.07/2.07, 1.50/1.50} {
        \draw[fixedLine]
            ([xshift=2pt,yshift={\yf cm-2.35cm}]A_fwd.east) --
            ([xshift=-2pt,yshift={\yb cm-2.35cm}]A_bwd.west);
    }

    \node[font=\scriptsize\sffamily\bfseries, text=black!70] at (2.40, 4.38)
        {Predefined Correspondence};

    \node[font=\scriptsize\sffamily, text=black!75, align=center] at (2.40, -0.30)
        {Assigned, one-to-one};
\end{scope}

% Subtle vertical divider
\draw[black!15, thin] (4.90, 0.35) -- (4.90, 5.10);

% =========================================================================
% PANEL B: Self-Assembled Routing (SAM)
% =========================================================================
\begin{scope}[xshift=5.35cm]
    \node[anchor=west, font=\small\sffamily\bfseries] at (0.0, 4.95)
        {B \quad Self-Assembled Routing (SAM)};

    % --- Stage 1: Random Many-to-Many Field ---
    \node[font=\scriptsize\sffamily\bfseries, text=black!70] at (2.15, 4.38)
        {Initial: Random Connectivity};

    \node[popBlock=fwdCol, minimum height=2.15cm] (B1_fwd) at (0.75, 2.35)
        {Forward\\stream};
    \node[popBlock=bwdCol, minimum height=2.85cm] (B1_bwd) at (3.55, 2.35)
        {Backward\\stream};

    \draw[driveArrow=fwdCol] (0.75, 0.35) -- (B1_fwd.south);
    \draw[driveArrow=fwdCol] (B1_fwd.north) -- (0.75, 4.15);
    \draw[driveArrow=bwdCol] (3.55, 4.20) -- (B1_bwd.north);
    \draw[driveArrow=bwdCol] (B1_bwd.south) -- (3.55, 0.35);

    % Soft random field with sparse crossing lines
    \fill[modCol!13]
        ([xshift=3pt,yshift=0.82cm]B1_fwd.east) --
        ([xshift=-3pt,yshift=1.15cm]B1_bwd.west) --
        ([xshift=-3pt,yshift=-1.15cm]B1_bwd.west) --
        ([xshift=3pt,yshift=-0.82cm]B1_fwd.east) -- cycle;
    \foreach \yf/\yb in {0.70/-0.90, 0.45/0.85, 0.00/-0.45, -0.55/0.45, -0.80/-0.95, -0.20/1.00} {
        \draw[modCol!45, thin, densely dashed, shorten <=2pt, shorten >=2pt]
            ([xshift=3pt,yshift={\yf cm}]B1_fwd.east) --
            ([xshift=-3pt,yshift={\yb cm}]B1_bwd.west);
    }

    \node[font=\scriptsize\sffamily, text=modCol!90!black, align=center] at (2.15, -0.30)
        {Unassigned, many-to-many};

    % --- Transition Arrow ---
    \draw[->, line width=2pt, black!55] (4.45, 2.35) -- (5.80, 2.35)
        node[midway, above=3pt, font=\scriptsize\sffamily\bfseries, text=black!80] {Local}
        node[midway, below=3pt, font=\scriptsize\sffamily\bfseries, text=black!80] {plasticity};

    % --- Stage 2: Aligned Cross-Stream Bridges ---
    \node[font=\scriptsize\sffamily\bfseries, text=black!70] at (8.10, 4.38)
        {Assembled: Aligned Bridges};

    \node[popBlock=fwdCol, minimum height=2.15cm] (B2_fwd) at (6.70, 2.35)
        {Forward\\stream};
    \node[popBlock=bwdCol, minimum height=2.85cm] (B2_bwd) at (9.50, 2.35)
        {Backward\\stream};

    \draw[driveArrow=fwdCol] (6.70, 0.35) -- (B2_fwd.south);
    \draw[driveArrow=fwdCol] (B2_fwd.north) -- (6.70, 4.15);
    \draw[driveArrow=bwdCol] (9.50, 4.20) -- (B2_bwd.north);
    \draw[driveArrow=bwdCol] (B2_bwd.south) -- (9.50, 0.35);

    % Coherent shaded field representing aligned cross-stream bridges
    \fill[modCol!18]
        ([xshift=3pt,yshift=0.82cm]B2_fwd.east) --
        ([xshift=-3pt,yshift=1.15cm]B2_bwd.west) --
        ([xshift=-3pt,yshift=-1.15cm]B2_bwd.west) --
        ([xshift=3pt,yshift=-0.82cm]B2_fwd.east) -- cycle;
    \fill[modCol!30]
        ([xshift=3pt,yshift=0.72cm]B2_fwd.east) --
        ([xshift=-3pt,yshift=1.02cm]B2_bwd.west) --
        ([xshift=-3pt,yshift=0.27cm]B2_bwd.west) --
        ([xshift=3pt,yshift=0.25cm]B2_fwd.east) -- cycle;
    \fill[modCol!25]
        ([xshift=3pt,yshift=0.08cm]B2_fwd.east) --
        ([xshift=-3pt,yshift=0.20cm]B2_bwd.west) --
        ([xshift=-3pt,yshift=-0.55cm]B2_bwd.west) --
        ([xshift=3pt,yshift=-0.50cm]B2_fwd.east) -- cycle;
    \fill[modCol!30]
        ([xshift=3pt,yshift=-0.58cm]B2_fwd.east) --
        ([xshift=-3pt,yshift=-0.70cm]B2_bwd.west) --
        ([xshift=-3pt,yshift=-1.15cm]B2_bwd.west) --
        ([xshift=3pt,yshift=-0.82cm]B2_fwd.east) -- cycle;

    \draw[<->, thick, dashed, modCol!95!black, shorten <=2pt, shorten >=2pt]
        ([xshift=3pt,yshift=0.48cm]B2_fwd.east) -- ([xshift=-3pt,yshift=0.68cm]B2_bwd.west);
    \draw[<->, thick, dashed, modCol!95!black, shorten <=2pt, shorten >=2pt]
        ([xshift=3pt,yshift=-0.05cm]B2_fwd.east) -- ([xshift=-3pt,yshift=-0.05cm]B2_bwd.west);
    \draw[<->, thick, dashed, modCol!95!black, shorten <=2pt, shorten >=2pt]
        ([xshift=3pt,yshift=-0.58cm]B2_fwd.east) -- ([xshift=-3pt,yshift=-0.78cm]B2_bwd.west);

    \node[font=\scriptsize\sffamily, text=modCol!90!black, align=center] at (8.10, -0.30)
        {Aligned cross-stream bridges};
\end{scope}

\end{tikzpicture}%
}
\caption{\textbf{Fixed versus self-assembled routing in biologically inspired learning.} \textbf{(A)} Standard feedback-alignment formulations relax exact weight symmetry between ascending inference and descending error streams (solid driving arrows) but presuppose a fixed correspondence between matched populations across the two pathways \cite{Lillicrap2016}. \textbf{(B)} In the Self-Assembling Motif (SAM), the forward and backward streams need not match in size and begin linked by unassigned, many-to-many modulatory connections (dashed field) \cite{Liao2026}. Through local plasticity, these connections organize into aligned cross-stream bridges. In stacked networks, this organization can support gradient-like learning and emergent preconditioning without requiring pre-wired correspondence.}
\label{fig:sam_conceptual}
\end{figure*}